\documentclass[lettersize,journal]{IEEEtran}
\usepackage{amsmath,amsfonts}
\usepackage{algorithmic}
\usepackage{algorithm}
\usepackage{array}
\usepackage{booktabs}
\usepackage[caption=false,font=normalsize,labelfont=sf,textfont=sf]{subfig}
\usepackage{textcomp}
\usepackage{stfloats}
\usepackage{url}
\usepackage{verbatim}
\usepackage{graphicx}
\usepackage{cite}
\usepackage{subfig}
\usepackage{adjustbox}

\begin{document}

\title{LiquidGazeNet: Towards Event-Based Near-Eye Gaze Estimation under Low-Light Conditions}

\author{Himanshu Kumar, and Sanjay Singh}




\maketitle

\begin{abstract}
Near-eye gaze estimation under low-light nighttime conditions is
challenging because the quality and effective temporal resolution of
conventional intensity frames can be reduced. This paper presents
LiquidGazeNet, an event-only gaze-estimation framework that combines
polarity-aware spike-latency encoding with a custom LTC-based
recurrent module. Positive- and negative-polarity events are
represented in separate channels, while exponentially decayed time
surfaces retain per-pixel temporal recency across successive event
windows. A shared convolutional encoder extracts spatial features from
each window, and a two-layer recurrent module composed of LTC-inspired
cells integrates them causally over short temporal sequences. The
network is trained using a hybrid objective that combines screen-space
coordinate regression with centre-referenced directional consistency.
The framework is evaluated on Gaze-FELL under cross-subject and
cross-scene protocols and is further assessed through
event-representation, temporal-module, window-duration, and
computational ablations. LiquidGazeNet achieves mean errors of
$3.75^{\circ}$ and $2.41^{\circ}$ under cross-subject and cross-scene
evaluation, respectively. The results indicate that polarity-aware
temporal-recency encoding, together with recurrent integration, is
effective for event-only gaze estimation under the evaluated low-light
nighttime conditions.
\end{abstract}


\section{Introduction}

Human gaze estimation has emerged as a critical sensing capability
for next-generation computing frameworks, spanning human--robot
interaction, extended reality (XR), and adaptive user interfaces
\cite{chukoskie2018quantifying,belardinelli2023gaze_intention,zhu2024methodology}.
In these emerging computing contexts, eye tracking enables
intent-aware interaction, foveated rendering, and
attention-driven resource allocation \cite{zhao2023ev,angelopoulos2020event}.
As gaze-sensing models move beyond controlled laboratory settings
towards wearable and nighttime applications, maintaining reliable
performance under reduced illumination becomes an important practical
challenge \cite{yin2024robust}.

Frame-based gaze estimation has achieved strong performance under
controlled conditions.
Appearance-based deep learning methods directly regress gaze from
eye or face images using convolutional and recurrent architectures
\cite{akinyelu2020cnn_survey,pathirana2022survey,cheng2024appearance,wang2021vision},
and calibration-aware networks, domain adaptation, and
transformer-based temporal modeling have further improved
robustness across subjects and environments
\cite{balim2023efe,ren2023fenet,zhou2020improved_itracker,
chen2020geddnet,tan2025harl,10794202}.
A fundamental limitation of frame-based sensors, however, is their
dependence on adequate photon flux: under the evaluated low-light conditions, the associated grayscale
observations exhibit reduced visibility and a low effective usable
rate, motion blur corrupts
spatial detail, and uniform-stride temporal models built on
frame sequences become unreliable
\cite{cheng2024appearance,wang2021vision}.
Infrared-based near-eye frameworks mitigate this through active
illumination but at the cost of constrained deployment geometry,
additional hardware, and reduced portability \cite{kim2019nvgaze}.
Low-light image enhancement can partially restore degraded frames
but introduces processing latency and reconstruction artifacts
that compromise the temporal fidelity required for precise gaze
trajectory estimation \cite{kim2020gaze_dark}.

Event cameras offer a fundamentally different computational sensing
paradigm.
Unlike frame-based sensors, event cameras asynchronously encode
per-pixel log-intensity changes with microsecond temporal resolution
and a dynamic range exceeding 120~dB, continuing to generate
informative signals even when conventional frames are nearly black
\cite{gallego2019survey}.
This sensing modality is relevant to gaze estimation under low-light
conditions: the asynchronous,
spike-driven nature of event data connects directly to computational
intelligence models that operate on irregular, time-varying inputs
\cite{iddrisu2023survey,zhao2023ev,angelopoulos2020event}.
Recent models have explored event-only gaze pipelines
\cite{li2024egaze,groenen2025gazescrnn} as well as hybrid
architectures that fuse event streams with conventional frames
\cite{zhao2024rgbez,chen2025exgaze,li2024modeling_state_shifting,
banerjee2024generalized_gaze_vector}.

\begin{figure}[t]
    \centering
    \includegraphics[width=1\linewidth]{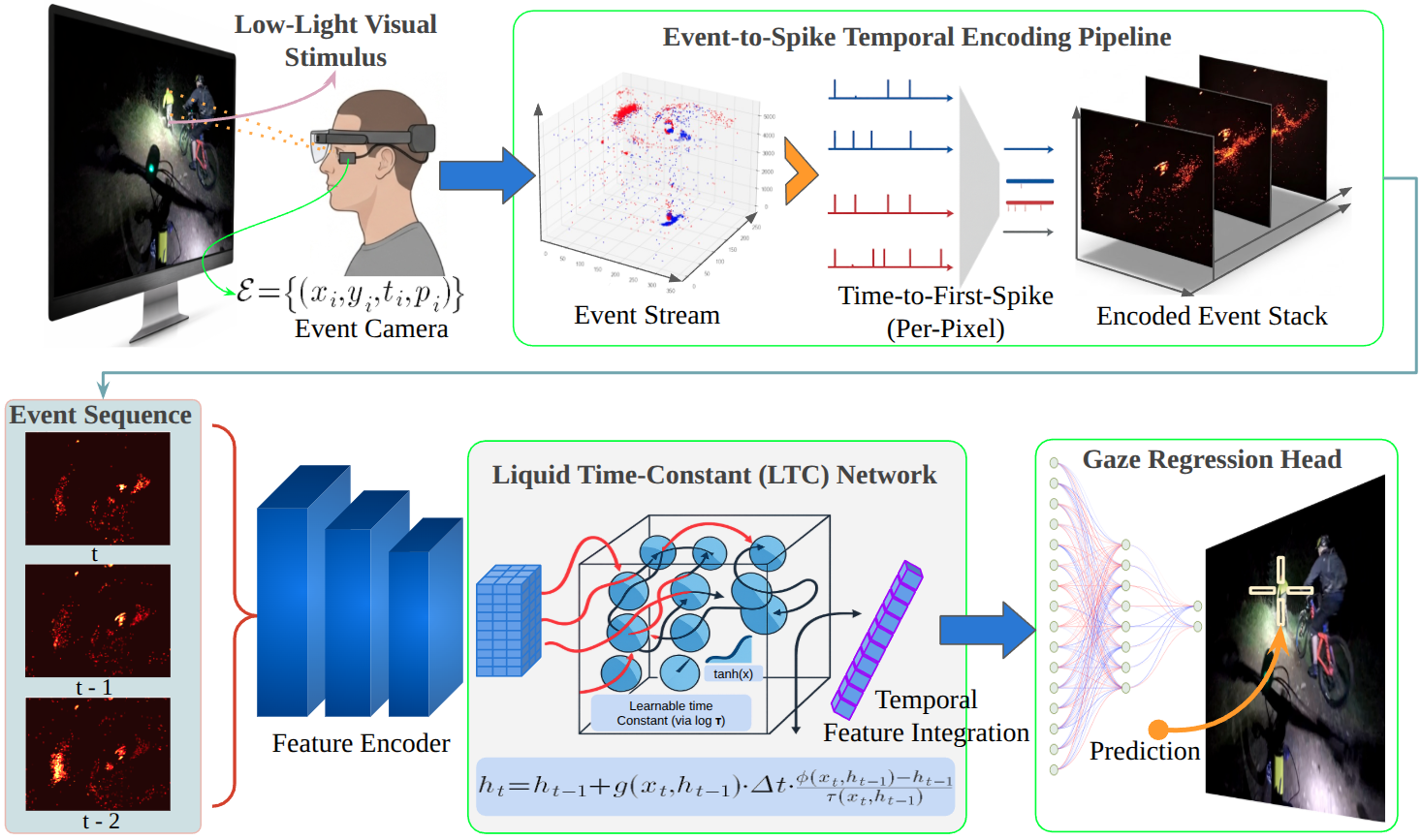}
    \caption{Overview of the proposed event-driven gaze-estimation
pipeline under display-only low-light illumination. A near-eye
DAVIS-346 sensor records asynchronous events, which are converted into
two-channel polarity-aware spike-latency maps using separate decayed
time surfaces for positive and negative events. Sequences of
$T=3$ consecutive 33~ms windows are processed by a shared convolutional
encoder and a two-layer LTC-based recurrent module. Each LTCi cell
combines an input-dependent candidate state, update gate, and positive
time constant through the gated residual update in
Equation~(\ref{eq:ltc_update}). The regression head outputs normalised
gaze coordinates $(\hat{x},\hat{y})\in[0,1]^2$.}
    \label{fig:overview}
\end{figure}

Under the display-only illumination represented by Gaze-FELL, the
associated grayscale observations exhibit reduced visibility and a low
effective usable rate. These conditions limit the spatial and temporal
information available to appearance-based gaze-estimation methods and
reduce the usefulness of hybrid event--frame processing. Existing
event-only gaze-estimation methods have mainly been evaluated under
controlled acquisition conditions, whereas prior work on Gaze-FELL has
used both event streams and grayscale guide frames. The performance
attainable from events alone on this dataset therefore remains
insufficiently examined.

A second challenge concerns temporal modelling. Although events are
asynchronous, learning-based pipelines commonly aggregate them into
fixed-duration representations before sequential processing. Successive
windows are uniformly spaced, but their visual content and event
activity can vary substantially with eye motion, illumination, and scene
content. Conventional recurrent models such as GRUs and LSTMs
\cite{8053243,10.1162/neco.1997.9.8.1735} provide effective discrete
temporal integration, while Liquid Time-Constant networks
\cite{hasani2021liquid} introduce input-dependent time constants that
modulate hidden-state evolution. Motivated by this principle, we develop
an LTC-inspired recurrent cell for fixed-window event representations.
The cell combines an input-dependent candidate state, a sigmoid update
gate, and a Softplus-parameterised positive time constant within a gated
Euler residual update.

This paper presents LiquidGazeNet, an event-only gaze-estimation
framework that combines polarity-aware spike-latency encoding with
LTC-inspired recurrent modelling. The framework is evaluated on
Gaze-FELL under cross-subject and cross-scene generalisation protocols
without test-time subject-specific adaptation. The main contributions
are:

\begin{enumerate}
    \item A polarity-aware spike-latency representation that preserves
    positive- and negative-polarity activity in separate channels while
    retaining per-pixel temporal recency across successive event
    windows.

    \item A custom LTC-based recurrent module composed of LTC-inspired cells that causally integrates features over $T=3$ event windows using a candidate state, update gate, and Softplus-parameterised positive time constant within a gated Euler residual update.

    \item A hybrid training objective combining screen-space coordinate
    regression with centre-referenced directional consistency.

    \item An event-only evaluation on Gaze-FELL under cross-subject and
    cross-scene protocols, supported by controlled ablations of the
    event representation, temporal module, window duration, and
    computational cost.
\end{enumerate}

\section{Related Work}

\subsection{Frame-Based and Low-Light Gaze Estimation}

Appearance-based deep learning methods for gaze estimation have
matured considerably, achieving strong performance through
CNN--RNN hybrids, attention mechanisms, temporal modeling, and
domain adaptation
\cite{akinyelu2020cnn_survey,pathirana2022survey,cheng2024appearance,
wang2021vision,ghosh2021automatic,zhou2020improved_itracker,
ren2023fenet,balim2023efe,chen2020geddnet,tan2025harl}.
These methods generally depend on sufficiently informative image
observations and approximately uniform frame acquisition, both of
which can deteriorate substantially under low-light nighttime conditions \cite{cheng2024appearance,wang2021vision}.
Active infrared near-eye architectures alleviate illumination limitations
\cite{kim2019nvgaze}, but require dedicated illumination hardware,
controlled sensor geometry, and additional power, which can restrict
their use in lightweight wearable architectures. Image-enhancement and reconstruction methods can partially recover
visual information from underexposed observations
\cite{kim2020gaze_dark}. However, enhancement introduces an additional
processing stage and may generate reconstruction artifacts that affect
the temporal and spatial cues required for precise gaze estimation.

Person-specific appearance and geometric differences remain another
important challenge in gaze estimation. Few-shot calibration methods
reduce subject-dependent prediction bias using a limited number of
labelled calibration samples \cite{park2019fewshot}, while gaze
redirection has been used to synthesise additional subject-specific
training observations \cite{yu2019gaze_redirection}. Cross-domain and
cross-person generalisation has also been addressed through
outlier-guided adaptation \cite{liu2021outlier_guided}, contrastive
regression \cite{wang2022contrastive_regression}, and source-free
uncertainty reduction \cite{cai2023sourcefree_gaze}. These studies
establish illumination variation and person-specific differences as
two important factors in gaze-estimation performance.

\subsection{Event-Based Eye Tracking and Gaze Estimation}

Event cameras asynchronously measure per-pixel log-intensity changes
with microsecond-scale temporal resolution and high dynamic range,
motivating their use for rapid eye-motion analysis
\cite{gallego2019survey,iddrisu2023survey}.
EV-Eye \cite{zhao2023ev} characterised the design space of
event-camera-based eye tracking and demonstrated the temporal
advantages of event sensing for high-frequency ocular motion.
The hybrid near-eye tracker of Angelopoulos et al.\
\cite{angelopoulos2020event} demonstrated kilohertz-rate pupil
tracking by combining asynchronous events with frame-based spatial
information under active infrared illumination.

Subsequent studies investigated purely event-based gaze-estimation
pipelines. E-Gaze \cite{li2024egaze} employed recurrent temporal
modeling and angular supervision, reporting a mean angular error of
$5.83^{\circ}$ on EV-Eye under uncalibrated cross-subject evaluation.
GazeSCRNN \cite{groenen2025gazescrnn} explored a spiking
convolutional recurrent architecture and reported $6.03^{\circ}$ on
the same benchmark. Recent benchmark studies have further compared
event-based eye-tracking methods in terms of estimation accuracy,
model complexity, and hardware-oriented efficiency
\cite{chen2025exgaze}. Hybrid networks such as EX-Gaze \cite{chen2025exgaze} and RGBE-Gaze
\cite{zhao2024rgbez} combine event streams with conventional frames
for high-frequency gaze tracking in extended-reality and remote-gaze
settings. Inference-time motion-aware refinement
\cite{bandara2025gaze_refinement} and local--global temporal
distillation \cite{li2024modeling_state_shifting} have additionally
been investigated to improve prediction coherence in event-based and
event--frame gaze estimation.

The Gaze-FELL dataset
\cite{banerjee2024generalized_gaze_vector} targets low-light gaze
estimation during the observation of egocentric nighttime scenes. Its
accompanying method combines event streams with grayscale guide frames
represented through temporal surfaces. Existing evaluations on
Gaze-FELL have therefore focused on hybrid event--frame processing,
leaving the performance of event-only gaze estimation on this dataset
insufficiently examined.

\subsection{Event Representations for Gaze Estimation}

Event representations range from simple accumulation images to
learned asynchronous embeddings. Count-based representations
accumulate positive and negative events over a fixed temporal window
but discard the relative event timestamps within that interval
\cite{gallego2019survey}. Time surfaces instead retain the timestamp
or recency of the latest event at each pixel, thereby preserving
local temporal information \cite{gallego2019survey}.

Discretised event volumes organise events over spatial and temporal
bins, retaining a coarse representation of their within-window
temporal distribution. Such representations have been used for
event-based motion estimation, including optical-flow, depth, and
egomotion prediction \cite{zhu2019unsupervised_event}. Event-to-image and event-to-video
reconstruction methods transform asynchronous measurements into
conventional intensity observations before applying frame-based
vision algorithms \cite{rebecq2019events_to_video}. Although such
reconstruction permits the use of established image-based networks,
it introduces a separate reconstruction stage and does not directly
exploit the original asynchronous representation.

Spike-latency coding \cite{li2024egaze} transforms event-recency
values into continuous latency measurements in which recently active
pixels are assigned smaller latency values. Compared with count
images, this representation retains information about event recency
while remaining compatible with standard convolutional operations. Together, these representations expose a trade-off between compact
spatial accumulation and preservation of temporal information.
Count-based maps retain event activity but discard event timing,
whereas time-surface and latency-based representations preserve
per-pixel recency while remaining compatible with convolutional
processing. Polarity-specific latency channels additionally retain the
sign of the underlying brightness changes.

\subsection{Continuous-Time and Neuromorphic Temporal Modeling}

Conventional recurrent architectures, including GRUs
\cite{8053243} and LSTMs
\cite{10.1162/neco.1997.9.8.1735}, process sequential observations through
discrete recurrent updates. When events are accumulated into
fixed-duration windows, the recurrent inputs are uniformly spaced in
time, but the information contained in each input can vary
substantially because event density depends on illumination, eye
motion, and scene activity.

Continuous-time sequence models provide an alternative formulation
in which hidden-state evolution is represented through differential
dynamics. Neural ODEs \cite{NEURIPS2018_69386f6b} introduced
continuous-depth latent-state evolution, while latent ODE models
extended this framework to irregularly sampled observations
\cite{rubanova2019latent_ode}. GRU-ODE-Bayes combines continuous
hidden-state evolution with gated observation updates
\cite{debrouwer2019gru_ode_bayes}, and neural controlled differential
equations treat the observation path as a continuous control signal
for the latent trajectory \cite{kidger2020neural_cde}. More recently,
continuous-time attention mechanisms have also been investigated for
irregular time-series modeling \cite{chen2023contiformer}. These formulations illustrate how latent-state dynamics can be adapted
to irregular or continuously evolving observations. Liquid Time-Constant networks
\cite{hasani2021liquid} introduce input-dependent time constants that
modulate the rate of hidden-state evolution. Each unit can therefore
adjust its effective temporal response according to the current input
and recurrent state. This property is relevant to event-based sensing,
where successive fixed-duration windows may contain markedly
different amounts of visual evidence.

Spiking neural networks constitute a complementary neuromorphic
approach in which information is propagated through discrete spikes.
They have been applied to event-based eye tracking and gaze estimation
\cite{groenen2025gazescrnn}. Liquid and spiking networks provide complementary approaches to
temporal modelling. Spiking architectures propagate information
through discrete events, whereas liquid networks use continuous-valued
hidden states whose evolution is modulated by learned time constants. Near-eye gaze sequences contain both rapid transitions and comparatively
persistent intervals, motivating temporal models that can vary their
state-update behaviour according to the current input. Liquid
Time-Constant networks provide one such formulation through
input-dependent time constants, while gated recurrent architectures
offer explicit control over state updates.

These complementary principles motivate an LTC-inspired recurrent
formulation for fixed-duration event representations. The formulation
combines an input-dependent positive time constant with candidate-state
and update-gate branches and applies a normalised Euler update at each
event window. Its detailed architecture and discrete recurrent
formulation are presented in Section~\ref{sec:methodology}.
\begin{figure}[t]
    \centering
    \includegraphics[width=1\linewidth]{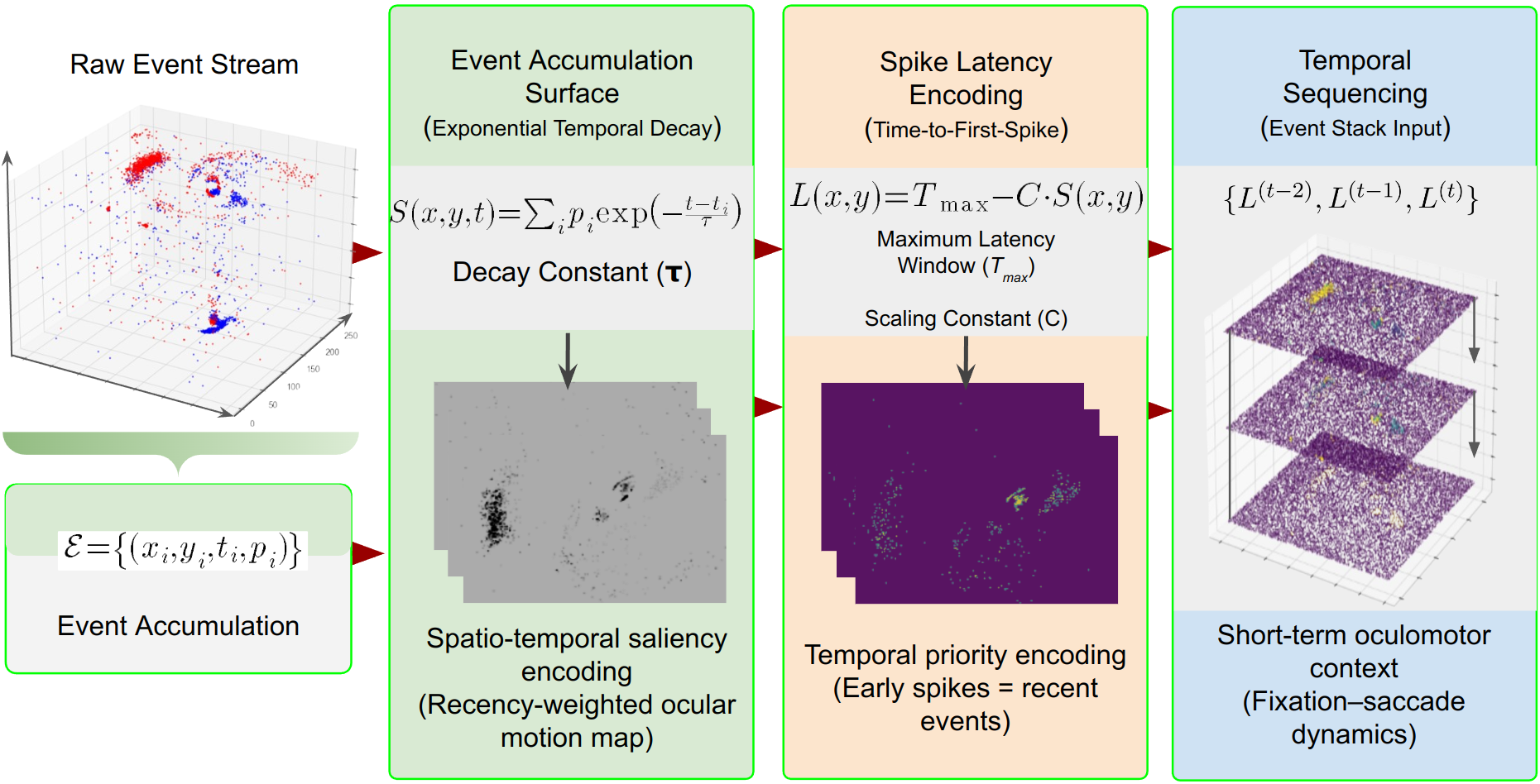}
    \caption{Polarity-aware spike-latency encoding. Separate exponentially
decayed time surfaces $\mathbf{S}^{+}$ and $\mathbf{S}^{-}$ retain the
recency of positive- and negative-polarity activity. Each surface is
mapped to a latency channel using
$\mathbf{L}^{\pm}=\operatorname{clip}
(T_{\max}-C\mathbf{S}^{\pm},0,T_{\max})$, where
$T_{\max}=30$~ms and $C=25$. Larger surface values therefore produce
smaller latency values. Three consecutive latency maps form the
temporal input sequence.}
    \label{fig:event_encoding}
\end{figure}

\section{Methodology}
\label{sec:methodology}

The proposed pipeline processes raw asynchronous event streams in a
causal manner through four sequential stages:
(i)~polarity-aware spike-latency encoding,
(ii)~shared convolutional spatial feature extraction,
(iii)~temporal integration through a two-layer LTCi recurrent module,
and (iv)~gaze regression.
Fig.~\ref{fig:overview} shows the complete pipeline.

An event camera records brightness changes as a stream
\begin{equation}
e_i = (x_i,\, y_i,\, t_i,\, p_i),
\end{equation}
where $(x_i, y_i)$ denotes pixel location, $t_i$ the timestamp,
and $p_i \in \{-1, +1\}$ the polarity of the brightness change.
The fundamental challenge is converting this continuous,
asynchronous, and irregularly distributed stream into a
representation amenable to gradient-based learning while
retaining polarity-specific activity and temporal recency across
successive windows.
\begin{figure*}[t]
    \centering
    \includegraphics[width=1\linewidth]{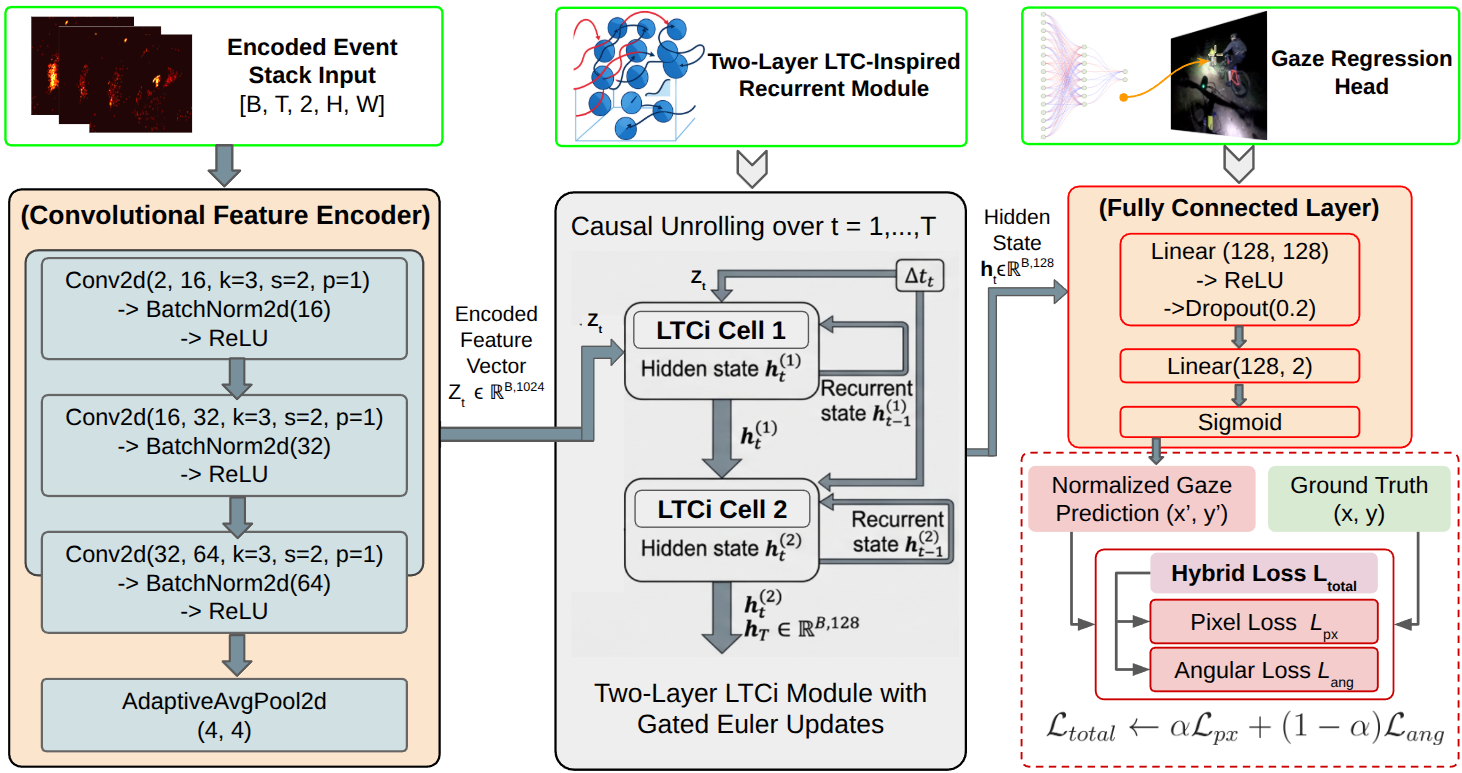}
    \caption{Architecture of LiquidGazeNet. Polarity-aware spike-latency
maps from $T=3$ event windows are encoded by a shared convolutional
backbone and projected to $128$-dimensional features. A two-layer LTCi
recurrent module then performs causal integration. Each LTCi cell
computes a candidate state, an update gate, and a
Softplus-parameterised positive time constant, which are combined
through a gated Euler update. The final hidden state
is decoded into normalised gaze coordinates and trained using the
hybrid screen-space and directional-consistency objective.}
    \label{fig:LiquidGazeNet}
\end{figure*}
\begin{algorithm}
\caption{Polarity-Aware Spike-Latency Event Encoding}
\label{alg:encoding}
\begin{algorithmic}
\STATE \textsc{ENCODE\_STREAM}$(\mathcal{E},\,\Delta T,\,\tau_s,\,
T_{\max},\,C)$
\STATE \textbf{Input:} Event stream
       $\mathcal{E}{=}\{(x_i,y_i,t_i,p_i)\}$,
       window duration $\Delta T$, decay $\tau_s$,
       maximum latency $T_{\max}$, scaling constant $C$
\STATE \textbf{Output:} Two-channel latency maps
       $\mathcal{L}{=}\{\mathbf{L}_1,\dots,\mathbf{L}_K\}$,\;
       $\mathbf{L}_k \in \mathbb{R}^{2 \times H \times W}$
\STATE $\mathbf{S}^{+},\mathbf{S}^{-} \leftarrow \mathbf{0}_{H\times W}$;\;
       $\mathcal{L}\leftarrow [\,]$
\FOR{each window $k$ of duration $\Delta T$}
    \STATE $\mathbf{S}^{+} \leftarrow
           \mathbf{S}^{+}\cdot\exp(-\Delta T/\tau_s)$
           \hfill\COMMENT{decay both surfaces}
    \STATE $\mathbf{S}^{-} \leftarrow
           \mathbf{S}^{-}\cdot\exp(-\Delta T/\tau_s)$
    \FOR{each event $(x_i,y_i,t_i,p_i)\in\mathcal{E}_k$}
        \IF{$p_i = +1$}
            \STATE $\mathbf{S}^{+}[y_i,x_i]\leftarrow 1$
        \ELSE
            \STATE $\mathbf{S}^{-}[y_i,x_i]\leftarrow 1$
        \ENDIF
    \ENDFOR
    \STATE $\mathbf{L}^{+}_k \leftarrow
           \textsc{clip}(T_{\max}-C\mathbf{S}^{+},\,0,\,T_{\max})$
    \STATE $\mathbf{L}^{-}_k \leftarrow
           \textsc{clip}(T_{\max}-C\mathbf{S}^{-},\,0,\,T_{\max})$
    \STATE $\mathbf{L}_k \leftarrow
           \textsc{stack}(\mathbf{L}^{+}_k,\,\mathbf{L}^{-}_k)$
    \STATE Append $\mathbf{L}_k$ to $\mathcal{L}$
\ENDFOR
\STATE \textbf{return} $\mathcal{L}$
\end{algorithmic}
\end{algorithm}

\subsection{Polarity-Aware Spike-Latency Encoding}
\label{subsec:encoding}

Two independent exponentially decayed time surfaces, one per
polarity, are maintained.
Between windows of duration $\Delta T$, both surfaces decay as
\begin{equation}
\mathbf{S}^{\pm}
\;\leftarrow\;
\mathbf{S}^{\pm} \cdot \exp\!\left(-\Delta T / \tau_s\right),
\end{equation}
with decay constant $\tau_s{=}$ 50ms.
Within each window, arriving events refresh the corresponding
surface at their pixel location to~1.
Each surface is then independently mapped to a continuous latency
code:
\begin{equation}
\mathbf{L}^{\pm}(x,y)
=
\operatorname{clip}\!\left(
T_{\max}-C\,\mathbf{S}^{\pm}(x,y),
\,0,\,T_{\max}
\right),
\label{eq:latency}
\end{equation}
where $T_{\max}=30$~ms denotes the maximum latency and $C=25$
controls the influence of the time-surface value on the encoded
latency. Larger surface values correspond to more recent activity and
therefore produce smaller latency values. The clipping operation keeps
the encoded values within $[0,T_{\max}]$. The two polarity-specific
maps are then stacked as separate channels:
\begin{equation}
\mathbf{L}_k
= \mathrm{stack}\!\left(\mathbf{L}^{+}_k,\, \mathbf{L}^{-}_k\right)
\in \mathbb{R}^{2 \times H \times W},
\label{eq:two_channel}
\end{equation}
yielding a two-channel tensor per window that retains
polarity-specific brightness-change information and per-pixel
temporal recency.

Consecutive maps are assembled into a causal sliding-window
sequence of length $T{=}3$:
\begin{equation}
\mathbf{X}_{\mathrm{seq}}
\in \mathbb{R}^{B \times 3 \times 2 \times H \times W},
\label{eq:input_shape}
\end{equation}
where $B$ is the batch size, $H{=}260$, $W{=}346$ (DAVIS-346
sensor dimensions).
The gaze label of the \emph{last} window in each sequence is
used as the training target, ensuring causal supervision without
future leakage.

The choice of $\Delta T{=}33$~ms is motivated by the temporal
scale of ocular dynamics rather than by sensor resolution.
Human saccades unfold over 20--200~ms, with fixation durations
spanning 150--500~ms \cite{gallego2019survey}.
A 33~ms window captures the onset and early trajectory of an
individual saccade within a single encoding step, while $T{=}3$
consecutive windows span ${\approx}100$~ms total context,
sufficient to observe a complete fast saccade or the transition
between two fixations.
Shorter windows may contain insufficient event activity, whereas
longer windows may reduce temporal specificity. The effect of
$\Delta T$ is evaluated empirically in the window-size study. The 33~ms window corresponds to an equivalent rate of approximately
30~Hz and provides the lowest error among the evaluated window
settings.

\subsection{Convolutional Feature Encoding}
\label{subsec:conv_encoder}

Each temporal slice of $\mathbf{X}_{\mathrm{seq}}$ is processed
independently by a shared convolutional encoder, applying the
same spatial transformation across all time steps and relegating
temporal modelling to the subsequent LTCi recurrent module. The encoder comprises three successive $3{\times}3$ convolutional
layers with stride~2 and channel widths $(16,32,64)$, each
followed by batch normalisation and ReLU activation.
The first layer accepts $C{=}2$ input channels corresponding to
the two polarity-latency maps:
\begin{equation}
\mathrm{Conv}_{1}:\;
\mathbb{R}^{2\times 260\times 346}
\;\to\;
\mathbb{R}^{16\times 130\times 173}.
\end{equation}
Adaptive average pooling to $4{\times}4$ at the encoder output
yields a $64{\times}4{\times}4{=}1024$-dimensional feature vector
$\mathbf{f}_t$ per time step. A lightweight projection layer
(Linear $\to$ ReLU $\to$ Dropout $p{=}0.1$) maps this feature to

\begin{equation}
\mathbf{x}_t
=
\operatorname{Proj}(\mathbf{f}_t)
\in \mathbb{R}^{D_{\mathrm{hidden}}},
\qquad
D_{\mathrm{hidden}}=128,
\end{equation}

which is supplied to the LTCi recurrent module.

Weight sharing across time steps reduces parameter count and
ensures that all windows are embedded in a common feature space,
so that temporal differences between consecutive windows reflect gaze dynamics rather than encoder drift.

\subsection{LTC-Based Recurrent Integration}
\label{subsec:ltc}
\emph{LTCi} denotes the proposed LTC-inspired recurrent
cell, which combines an input-dependent positive time constant with
candidate-state and update-gate branches in a gated Euler residual
update. Two stacked LTCi cells constitute the temporal module of
LiquidGazeNet.
\subsubsection{Architecture and Design Rationale}

The temporal component of LiquidGazeNet is a custom LTC-based recurrent
module rather than a direct implementation of the original LTC
architecture. It adopts the liquid-network principle of learned,
input-dependent time constants, but combines this mechanism with a
candidate-state branch and a sigmoid update gate in a residual Euler
formulation designed for fixed-window event representations. Each LTCi cell receives the projected feature $\mathbf{x}_t$
and the previous hidden state $\mathbf{h}_{t-1}$. From this joint context, it computes
(i) an input-dependent candidate state,
(ii) a sigmoid gate controlling the update magnitude, and
(iii) a neuron-wise positive time constant.
The resulting cell is referred to as LTC-inspired because its temporal
update is motivated by liquid time-constant dynamics, while its exact
gated formulation is specific to the proposed gaze-estimation
architecture. The module is evaluated over fixed-duration event windows with a
normalised step $\Delta t=1$. It should therefore be interpreted as an
LTC-based discretised recurrent model, not as a fully asynchronous
continuous-time network operating directly on individual event
timestamps.

\begin{figure}[t]
    \centering
    \includegraphics[width=1\linewidth]{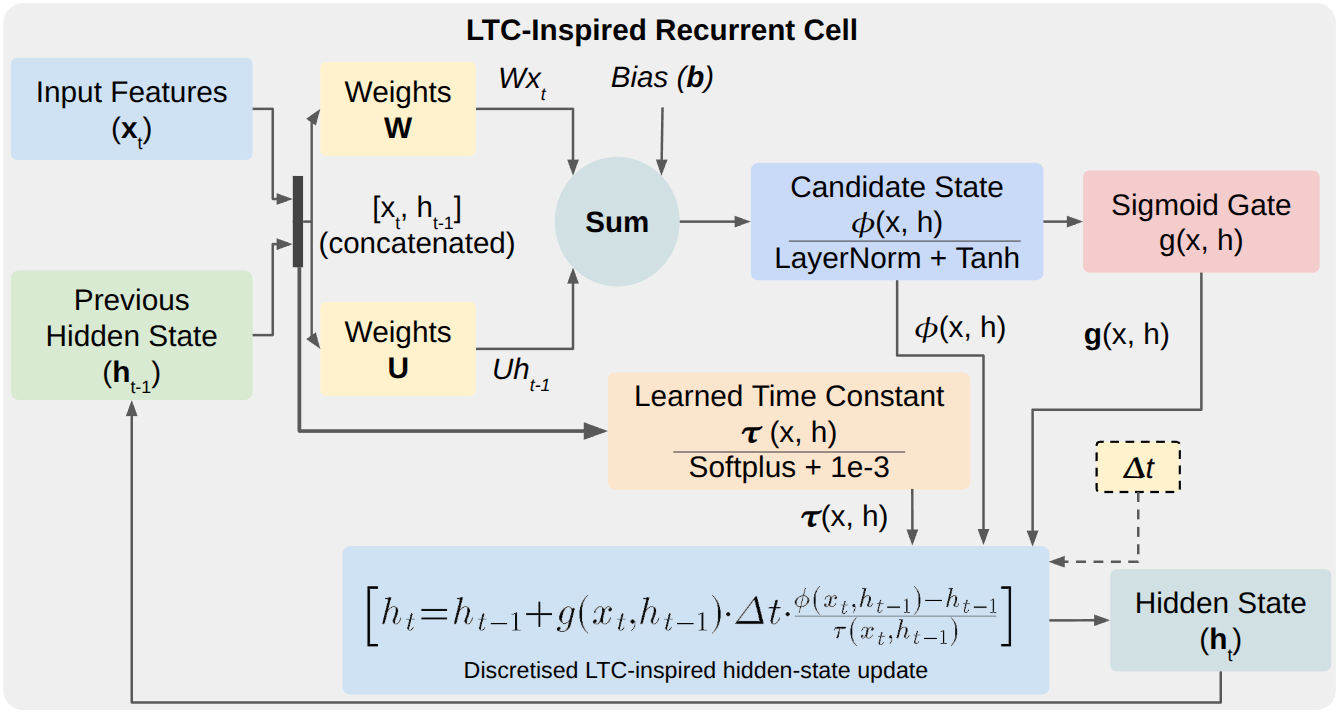}
    \caption{Proposed LTCi recurrent cell.
    The projected feature $\mathbf{x}_t$ and previous hidden state $\mathbf{h}_{t-1}$ are concatenated as $\mathbf{q}_t=[\mathbf{x}_t;\mathbf{h}_{t-1}]$, which feeds three parallel branches computing the candidate state
    $\boldsymbol{\phi}_t$ (Equation~\ref{eq:candidate}), the gate $g_t$
    (Equation~\ref{eq:gate}), and the time constant $\tau_t$
    (Equation~\ref{eq:tau}).
    The Softplus activation in the $\tau$ branch
    guarantees strict positivity for any unconstrained weight
    initialisation.
    The gated Euler update (Equation~\ref{eq:ltc_update}) combines
    all three to produce the next hidden state.}
    \label{fig:ltc_block}
\end{figure}

The proposed implementation uses a fixed-step update rather than an
adaptive Neural ODE solver \cite{NEURIPS2018_69386f6b}. This provides
a fixed computation graph for the $T=3$ event-window sequence.
Although the recurrent update is motivated by an ODE-style relaxation
process, it operates on temporally aggregated inputs and does not
directly integrate irregular event timestamps. The resulting
three-window context spans approximately 100~ms under the selected
33~ms window setting.

\subsubsection{LTC-Inspired Cell Formulation}
The implemented cell combines a liquid time-constant branch with
candidate-state and update-gate branches. The latter two components
define the gated recurrent adaptation used in LiquidGazeNet and are
not presented as an unchanged reproduction of the original LTC cell. Let $\mathbf{h}^{(l)}_t \in \mathbb{R}^{D_{\mathrm{hidden}}}$
denote the hidden state of LTCi layer $l\in\{1,2\}$ at step $t$,
and let $\mathbf{x}_t$ denote the projected encoder feature.
The concatenated input--state context is

\begin{equation}
\mathbf{q}_t
=
[\mathbf{x}_t;\mathbf{h}_{t-1}].
\label{eq:ltci_context}
\end{equation}

Three parallel branches compute:

\noindent\textbf{Candidate state:}
\begin{equation}
\boldsymbol{\phi}_t
=
\tanh\!\Bigl(
  \mathrm{LayerNorm}\bigl(
    \mathbf{W}\mathbf{x}_t
    + \mathbf{U}\mathbf{h}_{t-1}
    + \mathbf{b}
  \bigr)
\Bigr).
\label{eq:candidate}
\end{equation}

\noindent\textbf{Gate:}
\begin{equation}
g_t
=
\sigma\!\bigl(
  \mathbf{G}\mathbf{q}_t + \mathbf{c}
\bigr).
\label{eq:gate}
\end{equation}

\noindent\textbf{Time constant:}
\begin{equation}
\tau_t
=
\mathrm{Softplus}\!\bigl(
  \mathbf{T}\mathbf{q}_t + \mathbf{d}
\bigr)
+\epsilon,
\quad \epsilon = 10^{-3}.
\label{eq:tau}
\end{equation}

Strict positivity of $\tau_t$ is guaranteed by
$\mathrm{Softplus}(x) = \ln(1+e^{x}) > 0$ for all finite $x$,
together with the additive floor $\epsilon$.
Consequently, the weight matrix $\mathbf{T}$ and bias $\mathbf{d}$
can be initialised with any standard scheme (Kaiming uniform is
used) and remain fully unconstrained during optimisation:
positivity of $\tau_t$ is enforced structurally at
every forward pass, requiring no projection, clamping, or special
initialisation.

\subsubsection{Gated Euler Update and Unrolling}

The hidden state is updated via a gated Euler residual step:
\begin{equation}
\mathbf{h}_t
=
\mathbf{h}_{t-1}
+
g_t
\odot
\Delta t
\odot
\frac{
  \boldsymbol{\phi}_t - \mathbf{h}_{t-1}
}{
  \tau_t
},
\label{eq:ltc_update}
\end{equation}
where $\Delta t{=}1$ (normalised inter-window step) and $\odot$
denotes element-wise multiplication.
Equation~(\ref{eq:ltc_update}) can be interpreted as a gated,
first-order Euler-style discretisation of the relaxation dynamics
$\dot{\mathbf{h}}=g\odot(\boldsymbol{\phi}-\mathbf{h})/
\tau$. The sigmoid gate is an additional learned
modulation used in the proposed cell. Since $\Delta t=1$ and one update is applied per
fixed-duration event window, this equation defines a discretised
LTC-inspired recurrent update rather than direct continuous-time
integration of the raw event stream.

Two LTCi cells are stacked sequentially, with Layer~1 receiving
$\mathbf{x}_t$ and Layer~2 receiving $\mathbf{h}^{(1)}_t$.
Both layers are initialised with zero hidden states at the start
of each sequence.
The final state $\mathbf{h}^{(2)}_T$ serves as the
sequence-level representation of gaze dynamics.

\begin{algorithm}
\caption{Training Step}
\label{alg:training}
\begin{algorithmic}
\STATE \textsc{TRAIN\_STEP}$(\mathbf{X}_{\mathrm{seq}},\,
       \mathbf{Y}_{\mathrm{gt}},\,\alpha{=}0.8)$
\STATE \textbf{Input:}
  $\mathbf{X}_{\mathrm{seq}}\!\in\!
   \mathbb{R}^{B\times 3\times 2\times 260\times 346}$,\;
  $\mathbf{Y}_{\mathrm{gt}}\!\in\!\mathbb{R}^{B\times 2}$
  (label of last window)

\STATE $\mathbf{h}^{(1)}_0,\,\mathbf{h}^{(2)}_0
  \gets \mathbf{0}_{B\times D_{\mathrm{hidden}}}$

\FOR{$t = 1$ \textbf{to} $T{=}3$}
    \STATE $\mathbf{f}_t \gets
      \textsc{CNN}(
      \mathbf{X}_{\mathrm{seq}}[:,t,:,:,:])$
    \STATE $\mathbf{x}_t \gets
      \textsc{Proj}(\mathbf{f}_t)$
    \STATE $\mathbf{h}^{(1)}_t \gets
        \textsc{LTCiCell}_1(\mathbf{x}_t,\,
        \mathbf{h}^{(1)}_{t-1})$
    \STATE $\mathbf{h}^{(2)}_t \gets
      \textsc{LTCiCell}_2(\mathbf{h}^{(1)}_t,\,\mathbf{h}^{(2)}_{t-1})$
\ENDFOR

\STATE $\hat{\mathbf{g}} \gets
  \sigma\!\bigl(\mathrm{Linear}_2(
  \mathrm{Drop}_{0.2}(\mathrm{ReLU}(
  \mathrm{Linear}_{128}(\mathbf{h}^{(2)}_T))))\bigr)$

\STATE $\mathcal{L}_{\mathrm{px}} \gets
  \textsc{SmoothL1}(\hat{\mathbf{g}},\,\mathbf{Y}_{\mathrm{gt}})$
\STATE $\mathcal{L}_{\mathrm{ang}} \gets
  1 - \tilde{\mathbf{v}}_{\mathrm{pred}}^{\top}
       \tilde{\mathbf{v}}_{\mathrm{gt}}$
  \hfill\COMMENT{Equation~\ref{eq:angular_loss}}
\STATE \textbf{return}\;
  $\alpha\,\mathcal{L}_{\mathrm{px}} +
   (1{-}\alpha)\,\mathcal{L}_{\mathrm{ang}}$
\end{algorithmic}
\end{algorithm}

\subsection{Gaze Regression and Hybrid Training Objective}

The sequence-level representation $\mathbf{h}^{(2)}_T$ is decoded
to normalised gaze coordinates via a two-layer regression head:
\begin{equation}
\hat{\mathbf{g}}
= \sigma\!\bigl(
  \mathrm{Linear}_{2}\bigl(
    \mathrm{Drop}_{0.2}\bigl(
      \mathrm{ReLU}\bigl(
        \mathrm{Linear}_{128}(\mathbf{h}^{(2)}_T)
      \bigr)
    \bigr)
  \bigr)
\bigr) \in [0,1]^{2}.
\end{equation}

Training uses a hybrid pixel--angular loss:
\begin{equation}
\mathcal{L}
=
\alpha\,\mathcal{L}_{\mathrm{SmoothL1}}(\hat{\mathbf{g}},\,
\mathbf{g}_{\mathrm{gt}})
+
(1-\alpha)\,\mathcal{L}_{\mathrm{angular}},
\quad \alpha{=}0.8.
\label{eq:loss}
\end{equation}
The Smooth-$\ell_1$ term penalises screen-space deviations
robustly; the angular term enforces directional consistency via
centre-referenced gaze vectors.
Letting $\mathbf{c}=[W/2,H/2]$:
\begin{equation}
\mathbf{v}
= \mathbf{g} \odot [W,H] - \mathbf{c},
\quad
\tilde{\mathbf{v}}
= \mathbf{v}\,/\,(\|\mathbf{v}\|+\epsilon),
\end{equation}
\begin{equation}
\mathcal{L}_{\mathrm{angular}}
= 1 - \tilde{\mathbf{v}}_{\mathrm{pred}}^{\top}
        \tilde{\mathbf{v}}_{\mathrm{gt}}.
\label{eq:angular_loss}
\end{equation}
This formulation penalises directional error independently of
gaze vector magnitude, complementing the pixel-space penalty.
The $\alpha{=}0.8$ weighting reflects that screen-space accuracy
is the primary metric, with the angular term acting as a
regulariser on gaze direction.
The full training procedure is summarised in
Algorithm~\ref{alg:training}.

\section{Experimental Results}
\label{sec:experiments}

\subsection{Dataset and Experimental Setup}
\label{subsec:experimental_setup}

All experiments are conducted on Gaze-FELL
\cite{banerjee2024generalized_gaze_vector}, which contains DAVIS-346
near-eye recordings from five participants aged 25--35 years. The
participants viewed egocentric nighttime riding videos under
display-only illumination. The dataset contains four scene groups
comprising approximately 14 short clips per participant. Under these
conditions, the event stream remains responsive to local intensity
changes, whereas the associated grayscale frames are substantially
degraded and provide an effective usable rate of approximately
2--3~FPS. Gaze annotations are provided as screen coordinates
$(c_x,c_y)$ and are normalised to $[0,1]$ during training.

Positive- and negative-polarity events are encoded separately over
non-overlapping intervals of $\Delta T=33$~ms using
Algorithm~\ref{alg:encoding}, with $\tau_s=50$~ms,
$T_{\max}=30$~ms, and $C=25$. Each interval produces a two-channel
latency tensor of size $2\times260\times346$. Three consecutive tensors
are assembled into a sequence of length $T=3$, producing input batches
of size $[B,3,2,260,346]$. The gaze coordinate associated with the
final temporal step is used as the supervision target.

The dataset is partitioned at the complete-clip level before temporal
sequence construction. In each cross-subject fold, one participant is
held out for testing, and complete clips from the remaining
participants are divided into training and validation subsets using an
80/20 split with a fixed seed. Consequently, temporal sequences from
the same clip cannot occur across the training, validation, and test
partitions. In cross-scene evaluation, one complete scene group is
reserved for testing, while clips from the remaining scene groups are
used for model development. The calibrated setting uses a disjoint
subset of target-participant clips for lightweight adaptation and is
reported only as a personalised reference; calibration clips are
excluded from the corresponding evaluation set.

Models are trained using AdamW with an initial learning rate of
$5\times10^{-4}$ and weight decay $10^{-4}$. A ReduceLROnPlateau
scheduler is applied with patience 3 and reduction factor 0.5. Gradient
norms are clipped to 1.0. Training is performed for at most 100 epochs
with batch size 8. The recurrent hidden dimension is 128, the time-surface decay constant
is $\tau_s=50$~ms, the maximum latency is $T_{\max}=30$~ms, and the
latency scaling constant is $C=25$. Hidden states are initialised to
zero at the beginning of each independent sequence. Model selection is
based on validation performance, and the best validation checkpoint is
used for testing. All experiments are performed on an NVIDIA GeForce
RTX~3060 GPU.

The primary measures are mean pixel error, median pixel error, and
screen-centred directional error. The latter is calculated as the
angular separation between predicted and reference two-dimensional
vectors originating at the screen centre. It should therefore not be
interpreted as a calibrated three-dimensional gaze-ray error. The
coefficient of determination $R^2$ and Pearson correlation are also
reported for the horizontal and vertical gaze coordinates. Validation
mean average precision is computed over pixel-error thresholds
$\{20,40,60,80,100\}$~px.

\begin{figure}[t]
    \centering
    \includegraphics[width=1\linewidth]{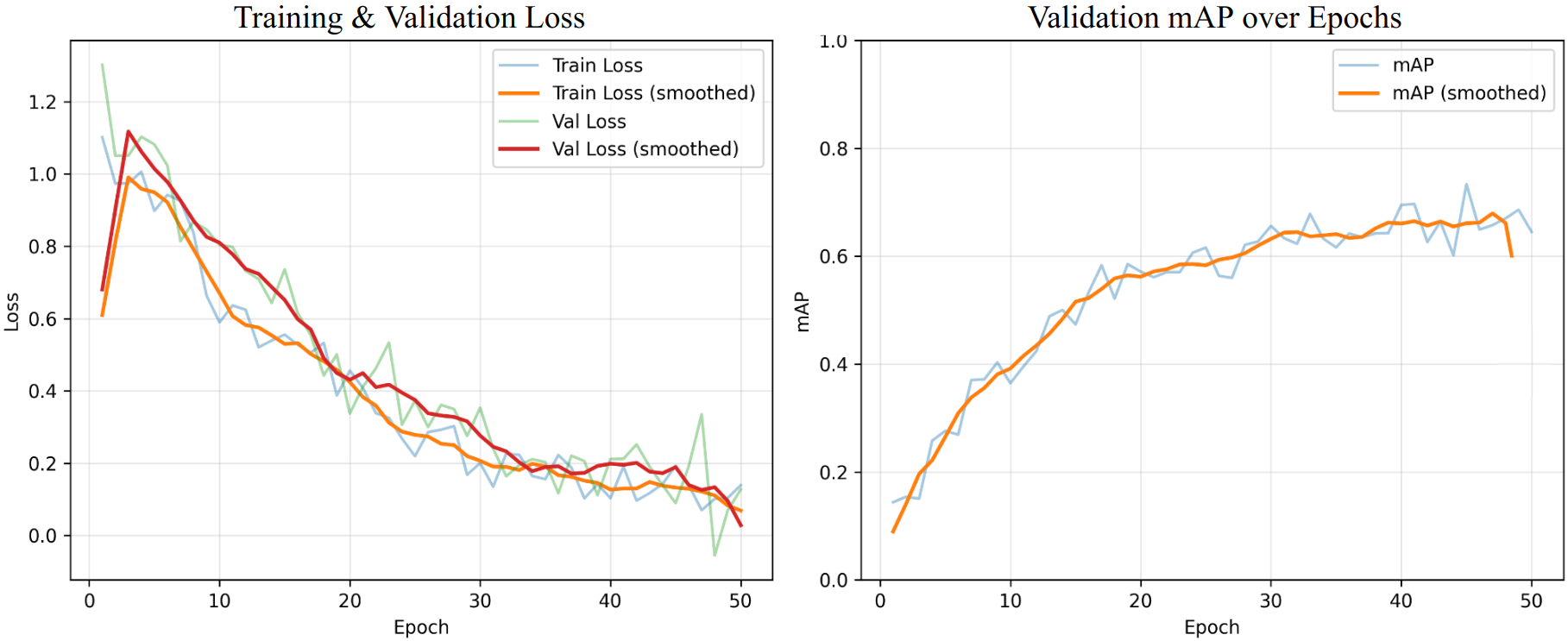}
    \caption{Training dynamics over the first 50 epochs. The left panel
    shows the raw and smoothed training and validation losses, while the
    right panel shows the raw and smoothed validation mean average
    precision computed over pixel-error thresholds
    $\{20,40,60,80,100\}$~px.}
    \label{fig:training_dynamics}
\end{figure}

\subsection{Gaze Estimation Performance}
\label{subsec:gaze_performance}

Figure~\ref{fig:training_dynamics} illustrates the optimisation
behaviour over the first 50 epochs. The training and validation losses
decrease with a similar overall trend, while validation mAP increases
and begins to stabilise during the later epochs shown. Table~\ref{tab:protocol_perf} summarises performance under the three
evaluation settings. LiquidGazeNet obtains a mean directional error of
$3.75^{\circ}$ under cross-subject evaluation and $2.41^{\circ}$ under
cross-scene evaluation. Because the cross-scene setting retains data
from the same participant during model development, its lower error is
consistent with participant variation being a more demanding factor
than variation among the evaluated scene groups. The calibrated result
of $1.43^{\circ}$ is reported only as a personalised reference and is
not directly comparable with the uncalibrated cross-subject setting.

\begin{table}[htbp]
\centering
\caption{Gaze-estimation performance under the evaluated cross-subject,
cross-scene, and calibrated protocols.}
\label{tab:protocol_perf}
\small
\resizebox{0.6\linewidth}{!}{%
\begin{tabular}{lc}
\toprule
\textbf{Evaluation protocol} & \textbf{Mean directional error} \\
\midrule
Cross-subject generalisation & $3.75^{\circ}$ \\
Cross-scene generalisation   & $2.41^{\circ}$ \\
Calibrated reference         & $1.43^{\circ}$ \\
\bottomrule
\end{tabular}}
\end{table}

Table~\ref{tab:cross_scene_participant} reports participant-wise
performance under the cross-scene setting. Mean directional error
ranges from $2.30^{\circ}$ to $2.55^{\circ}$. The corresponding $R^2$
and Pearson values show consistent agreement between predicted and
reference screen coordinates across the evaluated participants.

\begin{table}[htbp]
\centering
\caption{Per-participant gaze-estimation performance under the
cross-scene evaluation protocol. For each participant, one complete
scene group is held out for testing without test-time adaptation.}
\label{tab:cross_scene_participant}
\small
\resizebox{0.8\linewidth}{!}{%
\begin{tabular}{lcccc}
\toprule
\textbf{User} & \textbf{Mean px} & \textbf{Med.\ px} &
\textbf{Dir.} & \textbf{$R^{2}$ (X,Y) / Pearson (X,Y)} \\
\midrule
1 & 31.2 & 21.0 & 2.55$^{\circ}$ & (0.93,\,0.83) / (0.96,\,0.91) \\
2 & 29.4 & 20.1 & 2.43$^{\circ}$ & (0.94,\,0.85) / (0.97,\,0.92) \\
3 & 27.8 & 19.3 & 2.30$^{\circ}$ & (0.95,\,0.87) / (0.97,\,0.93) \\
4 & 30.1 & 20.6 & 2.47$^{\circ}$ & (0.94,\,0.84) / (0.96,\,0.91) \\
5 & 28.5 & 19.8 & 2.31$^{\circ}$ & (0.95,\,0.88) / (0.98,\,0.94) \\
\midrule
\textbf{Mean} & \textbf{29.4} & \textbf{20.2} &
\textbf{2.41$^{\circ}$} & \textbf{(0.94,\,0.85) / (0.97,\,0.92)} \\
\bottomrule
\end{tabular}}
\end{table}

\begin{figure}[!htbp]
\centering
\includegraphics[width=1\linewidth]{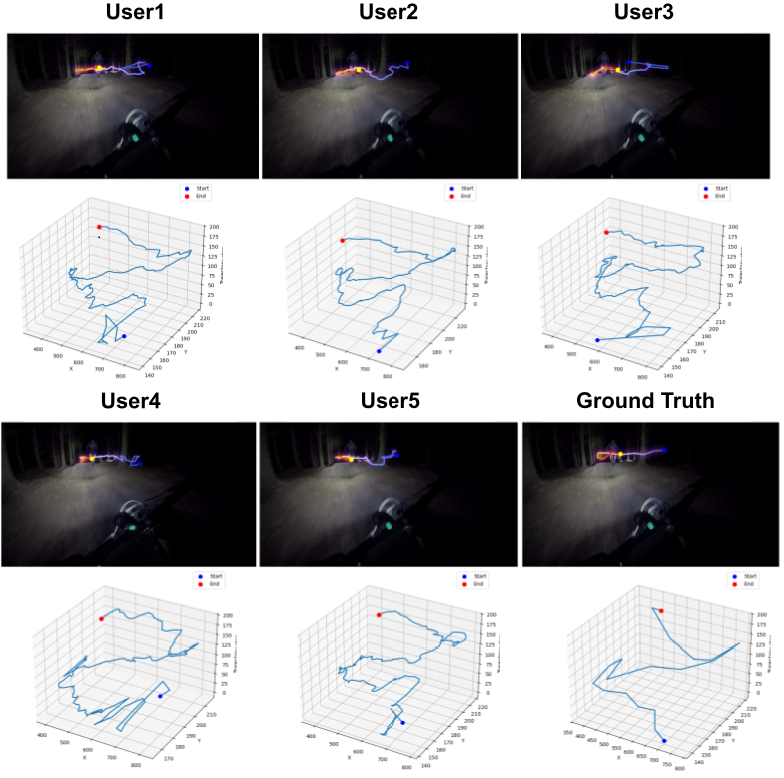}
\caption{Predicted and reference gaze trajectories for one representative
Gaze-FELL clip. The upper panels show the gaze path overlaid on the
video frame for each user and for the reference annotation. The lower
panels show the corresponding trajectories in $(x,y,t)$ space, where
the temporal axis indicates progression through the clip. Blue and red
markers denote the start and end of each trajectory, respectively.}
\label{fig:single_clip_trajectories}
\end{figure}

\begin{figure}[!htbp]
\centering
\includegraphics[width=1\linewidth]{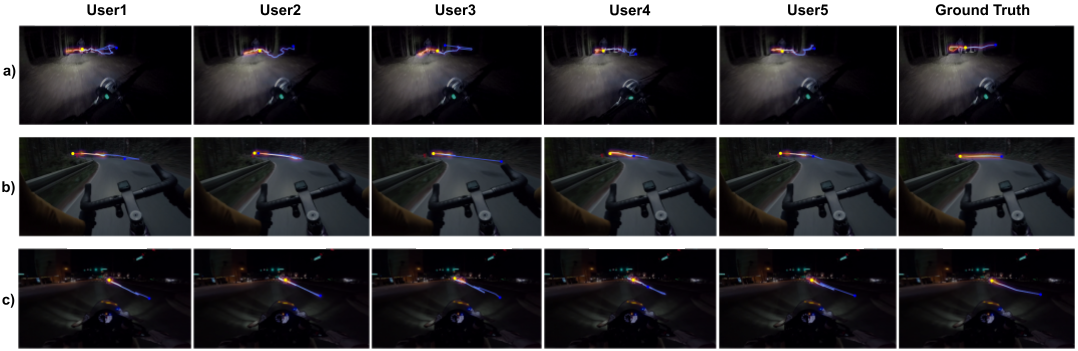}
\caption{Predicted and reference gaze trajectories for three representative
Gaze-FELL clips. Rows (a)--(c) correspond to different video samples,
while the columns show predictions for Users 1--5 and the corresponding
reference trajectory. The colour transition from blue to red indicates
temporal progression from the beginning to the end of each trajectory.}
\label{fig:trajectory}
\end{figure}

Figure~\ref{fig:single_clip_trajectories} presents the predicted and
reference gaze trajectories for one representative clip across all five
users. The frame overlays show the spatial gaze paths, while the
three-dimensional $(x,y,t)$ plots illustrate their temporal evolution.
Figure~\ref{fig:trajectory} provides additional qualitative examples
from three different clips. Across these examples, the predictions
generally recover the principal gaze regions and follow the overall
temporal progression of the reference trajectories, although local
deviations remain visible.

Table~\ref{tab:window_sensitivity} evaluates the event-accumulation
interval. The 33~ms interval obtains the lowest mean directional error
among the tested settings. Performance decreases for both shorter and
longer intervals within the evaluated range, suggesting a trade-off
between the amount of event activity represented in each map and its
temporal resolution. A direct event-density analysis would be required
to determine the precise cause of this trend.

\begin{table}[t]
\centering
\caption{Cross-subject mean directional error for different
event-window durations $\Delta T$, with all other experimental
settings fixed.}
\label{tab:window_sensitivity}
\small
\resizebox{0.6\linewidth}{!}{%
\begin{tabular}{lccccc}
\toprule
$\Delta T$ (ms) & 10 & 20 & \textbf{33} & 50 & 66 \\
\midrule
Mean directional error ($^{\circ}$) & 5.12 & 4.31 & \textbf{3.75} & 4.08 & 4.67 \\
\bottomrule
\end{tabular}}
\end{table}

Table~\ref{tab:efficiency} compares LTCi, GRU, and LSTM configurations
sharing the same encoder, projection layer, and regression head. The
LTCi model contains 369,522 parameters and has a measured latency of
2.552~ms per batch of eight sequences, compared with 2.008~ms for GRU
and 1.980~ms for LSTM. Peak memory and GFLOPs are similar across the
three configurations, whereas LTCi introduces a moderate latency and
throughput overhead. These GPU measurements characterise the evaluated
implementation and do not substitute for device-specific embedded
latency or power measurements.

\begin{table}[t]
\centering
\caption{Computational comparison of the GRU, LSTM, and LTCi temporal
modules under a shared convolutional encoder. Measurements are reported
for input size $[B,3,2,260,346]$ with batch size $B=8$ on an NVIDIA
GeForce RTX 3060.}
\label{tab:efficiency}
\small
\resizebox{0.6\linewidth}{!}{%
\begin{tabular}{lccc}
\toprule
\textbf{Metric} & \textbf{GRU} & \textbf{LSTM} & \textbf{LTCi} \\
\midrule
Parameters          & 369{,}778 & 435{,}826 & 369{,}522 \\
Model size (MB)      & 1.411     & 1.663     & 1.410 \\
Latency (ms/batch)   & 2.008     & 1.980     & 2.552 \\
Peak memory (MB)     & 91.90     & 92.68     & 91.90 \\
Throughput (batch/s) & 497.92    & 505.14    & 391.78 \\
GFLOPs/forward       & 2.862     & 2.866     & 2.872 \\
\bottomrule
\end{tabular}}
\end{table}

\subsection{Comparative Analysis}

Table~\ref{tab:cross_dataset_comparison} places LiquidGazeNet alongside
representative gaze-estimation methods spanning RGB, infrared, hybrid
event--frame, and event-only sensing. The comparison highlights the
different acquisition settings in which these methods are evaluated,
including natural RGB scenes, controlled near-eye infrared setups, and
display-only nighttime illumination. Because the reported results are
obtained on different datasets and under different calibration and
evaluation protocols, the table is used to position LiquidGazeNet within
the broader gaze-estimation literature rather than to establish a strict
numerical ranking.

To provide an additional event-based reference, LiquidGazeNet is also
evaluated on EV-Eye using a stateful configuration. The EV-Eye experiment
is reported separately from the Gaze-FELL results because the two datasets
differ in acquisition conditions, calibration assumptions, and error
definitions. Table~\ref{tab:eveye_comparison} compares the resulting
event-only configuration with representative EV-Eye methods, including
models that use frame initialisation and subject-specific calibration.
Within this setting, LiquidGazeNet-Stateful achieves a mean angular error
of $5.09^{\circ}$ with approximately $0.37$~M parameters, providing an
additional indication of the model's applicability beyond Gaze-FELL.
However, DoD and MAE remain distinct error measures and should be
interpreted according to the corresponding evaluation protocol.

\begin{table}[t]
\centering
\caption{Comparison of representative gaze-estimation methods under
different sensing modalities, datasets, and acquisition conditions.}
\label{tab:cross_dataset_comparison}
\small
\resizebox{\columnwidth}{!}{%
\begin{tabular}{lllc}
\toprule
\textbf{Method} & \textbf{Modality / dataset} &
\textbf{Acquisition condition} & \textbf{Reported error} \\
\midrule
ResNet-14 baseline~\cite{9642966} & RGB face / MPIIFaceGaze & Natural RGB & $4.71^{\circ}$ \\
FE-Net~\cite{ren2023fenet} & RGB face and eyes / MPIIFaceGaze & Natural RGB & $3.66^{\circ}$ \\
NVGaze~\cite{kim2019nvgaze} & Active IR / NVGaze & Controlled near-eye IR & $3.51^{\circ}$ \\
E-Gaze~\cite{li2024egaze} & Event only / EBVEYE & Controlled near-eye & $5.83^{\circ}$ \\
\textbf{LiquidGazeNet} & Event only / Gaze-FELL & Display-only nighttime illumination & $3.75^{\circ}$ \\
\bottomrule
\end{tabular}}
\end{table}

\begin{table}[t]
\centering
\caption{Comparison of representative event-based gaze-estimation
methods on the EV-Eye dataset under their reported input and
calibration settings. DoD denotes difference of direction, and MAE
denotes mean angular error.}
\label{tab:eveye_comparison}
\small
\resizebox{\columnwidth}{!}{%
\begin{tabular}{lllcc}
\toprule
\textbf{Method} & \textbf{Input} & \textbf{Calibration} &
\textbf{Error} & \textbf{Parameters} \\
\midrule
EVBEYE~\cite{angelopoulos2020event} & Events + frame initialisation & Subject-specific & $9.72^{\circ}$ DoD & N/A \\
EV-Eye benchmark~\cite{zhao2023ev} & Events + frame initialisation & Subject-specific & $4.71^{\circ}$ DoD & 17.27 M\\
GazeSCRNN~\cite{groenen2025gazescrnn} & Events only & None & $6.03^{\circ}$ MAE & 2.30 M \\
\textbf{LiquidGazeNet-Stateful} & Events only & None & $5.09^{\circ}$ MAE & 0.37 M \\
\bottomrule
\end{tabular}}
\vspace{2pt}
\begin{minipage}{0.98\columnwidth}
\
\end{minipage}
\end{table}

\subsection{Ablation Study}
\label{subsec:ablation}

Table~\ref{tab:ablation} evaluates the event representation and
temporal module under the same cross-subject protocol. All
configurations use two polarity channels, $T=3$, and
$\Delta T=33$~ms. The encoder, projection layer, regression head,
training schedule, and evaluation protocol are held fixed.

\begin{table}[h]
\centering
\caption{Effect of the event representation and temporal module on
cross-subject gaze-estimation performance.}
\label{tab:ablation}
\small
\resizebox{0.6\linewidth}{!}{%
\begin{tabular}{lc}
\toprule
\textbf{Configuration} & \textbf{Mean directional error} \\
\midrule
Binned events + CNN only & $8.07^{\circ}$ \\
Binned events + GRU      & $8.68^{\circ}$ \\
Binned events + LSTM     & $8.86^{\circ}$ \\
Binned events + LTCi     & $8.63^{\circ}$ \\
\midrule
Spike latency + CNN only & $4.78^{\circ}$ \\
Spike latency + GRU      & $3.93^{\circ}$ \\
Spike latency + LSTM     & $4.55^{\circ}$ \\
\textbf{Spike latency + LTCi} & $\mathbf{3.75^{\circ}}$ \\
\bottomrule
\end{tabular}}
\end{table}

The event representation produces the largest performance difference.
Replacing binned polarity counts with spike-latency maps reduces the
mean error by $3.85^{\circ}$--$4.31^{\circ}$ across the evaluated
temporal modules. Even without recurrent integration, the spike-latency CNN achieves $4.78^{\circ}$ and outperforms all recurrent
configurations based on binned counts. This indicates that per-pixel
temporal recency retains information that is not captured by the count maps used in this comparison. Temporal integration provides a further improvement for the
spike-latency representation. The proposed LTCi configuration reduces
the error from $4.78^{\circ}$ for the CNN-only model to
$3.75^{\circ}$, an absolute difference of $1.03^{\circ}$. Among the
evaluated recurrent alternatives, LTCi obtains the lowest mean error
under the shared encoder, training schedule, and evaluation protocol.
The larger LSTM configuration does not improve on GRU or LTCi,
indicating that parameter count alone does not explain the observed
performance. These results support the complete LTC-inspired recurrent formulation,
which combines candidate-state, update-gate, and input-dependent
positive time-constant branches. The observed gain therefore reflects
the complete LTCi formulation rather than any single component in
isolation.

\section{Conclusion}
\label{sec:conclusion}

This work presented LiquidGazeNet, an event-only framework for
near-eye gaze estimation under the display-only nighttime illumination
conditions represented by Gaze-FELL. The method combines
polarity-aware spike-latency encoding with a two-layer LTCi recurrent
module composed of LTC-inspired cells. The module integrates features
across $T=3$ consecutive event windows through a gated residual update
with input-dependent positive time constants. Positive- and
negative-polarity events are retained in separate channels, while the
latency representation preserves per-pixel temporal recency across
windows.

LiquidGazeNet achieved mean errors of $3.75^{\circ}$ under
cross-subject evaluation and $2.41^{\circ}$ under cross-scene
evaluation without subject-specific adaptation at test time. The
ablation results show that the largest improvement arises from the
spike-latency representation, which consistently outperforms
polarity-separated event-count maps across the evaluated temporal
modules. Recurrent integration provides an additional gain over the
single-window configuration, and the LTCi model obtains the lowest mean
error among the tested recurrent alternatives. The window-size study
further identifies $\Delta T=33$~ms with $T=3$ as the best-performing
tested configuration. The resulting model contains approximately
$0.37$~M parameters and operates without grayscale guide frames during
inference. The evaluation is limited to five participants, four scene groups,
a single event sensor, and one near-eye acquisition configuration.
Future work should evaluate the method on larger and more diverse
event-based gaze datasets, investigate temporal updates that explicitly
incorporate elapsed time, and assess deployment on near-eye computing
platforms.

\bibliographystyle{IEEEtran}
\bibliography{Reference.bib}

\vfill

\end{document}